\documentclass[8pt, twocolumn]{article}
\usepackage{mathptmx} 
\usepackage{parskip} 
\usepackage[left=0.75in,right=0.75in,top=1in,bottom=1in]{geometry}
\usepackage{amsmath, amssymb, amsfonts}
\usepackage{graphicx, wrapfig, array, multirow, rotating}
\usepackage[font=footnotesize,labelfont=bf]{caption}
\usepackage{booktabs} 
\usepackage[bottom]{footmisc}
\usepackage[rightcaption]{sidecap}

\usepackage [english]{babel}
\usepackage [autostyle, english = american]{csquotes}
\MakeOuterQuote{"}

\usepackage[hidelinks]{hyperref}
\usepackage[sortcites,
            backend=biber,
            giveninits=true, 
            hyperref=true]{biblatex}

\DeclareNameAlias{author}{family-given}
 
\begin{document}
\title{Modeling the Evolution of Retina Neural Network}
\author{%
\begin{tabular}{c} Ziyi Gong \\ University of Pittsburgh \\ zig9@pitt.edu \end{tabular} \and
\begin{tabular}{c} Paul Munro \\ University of Pittsburgh \\ pwm@pitt.edu \end{tabular} }
\date{}
\maketitle

    
    
    
    

\begin{abstract}
    \noindent
    Vital to primary visual processing, retinal circuitry shows many similar structures across a very broad array of species, both vertebrate and non-vertebrate, especially functional components such as lateral inhibition. This surprisingly conservative pattern raises a question of how evolution leads to it, and whether there is any alternative that can also prompt helpful preprocessing. Here we design a method using genetic algorithm that, with many degrees of freedom, leads to architectures whose functions are similar to biological retina, as well as effective alternatives that are different in structures and functions. We compare this model to natural evolution and discuss how our framework can come into goal-driven search and sustainable enhancement of neural network models in machine learning.
\end{abstract}


\section{Introduction}
Retina serves as the first visual processing unit in animals' visual systems. Not only does it transduce patterns of light energy light into spike trains, but in the process, it performs preprocesses the information to augment functions of perception and cognition, such as object identification. 

There have been a handful of studies on the evolution of morphology and genetics related to retina, and several theories are proposed. Classification of the available fossil records and current animals yields 3 groups: no 'retina', where ciliary photoreceptors give rise to axons directly; 2-layered retina, where photoreceptors make contact with retina ganglion cells (RGCs), spiking neurons that project to higher processing units in the brain; and 3-layered retina, where the additional layer between photoreceptors and RGCs consist of bipolar cells (BPs) and important regulatory interneurons such as horizontal cells (HCs) and amacrine cells (ACs). The evolution is believed to have progressed in the order depicted above, with hypothesized changes such as duplication and mutation of photoreceptors that produced ON-BPs, transduction of glial cells that led to HCs and ACs, mutation of ligand-gated channels that split OFF-BPs from ON-BPs, etc. \cite{evo_lamb}. Our model does not consider "no 'retina'" class and assumes that the evolutionary process starts at the time when a 2-layered retinal architecture had not yet stabilized.

In jawed vertebrates, the 3-layered retina, also the most complex ones, can be viewed as the composition of several functional components that together enable processing, such as the lateral inhibition facilitated by HCs, and center-surround arrangement of ON/OFF BPs. 

\begin{figure}
    \centering
    \includegraphics{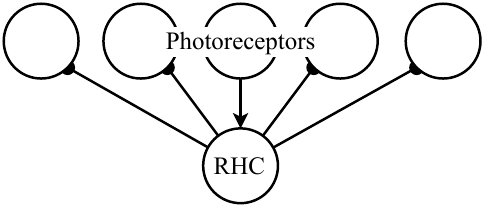}\\(a)\vspace{3mm}\\
    \includegraphics{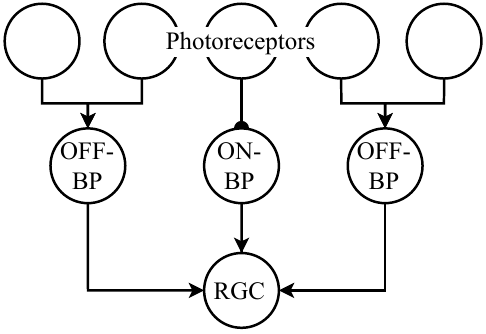}\\(b)\vspace{3mm}\\
    \caption{Arrow: excitatory connections; hemisphere: inhibitory connections. (a) Retinal lateral inhibition. (b) ON-center, OFF-surround structure.}
\end{figure}

Lateral inhibition, in which a group of cells suppresses the activities of their surrounding cells, firstly discovered in the eyes of \textit{Limulus} \cite{limulus}, has been found in various systems of multi-cellular organisms \cite{sensation}. Primary visual processing of visual animals, from insects to mammals, involves lateral inhibition to enhance sharpness and modify color discrepancy of visual scenes \cite{sensation}. In mammals, HCs, interneurons that play a critical role in primary retina lateral inhibition, receive inputs from surrounding photoreceptors and provides negative feedback to the nearest photoreceptors (Fig. 1a), facilitating the formation of center-surround receptive fields \cite{lateralinhib}. 

The center-surround arrangement contains either ON- or OFF-BPs responsible for the center, and another type for the surround (Fig. 1b). ON-BPs are hyperpolarized when the light is insufficient, and that is the reverse for OFF-BPs; both of them activate the RGCs. Thus, an ON-center, OFF-surround arrangement will allow the corresponding RGC to fire the most when the center is illuminated while the surround is not. It furthermore leads to the center-surround receptive field of retina ganglion cells (RGCs), the output neurons of the retina. The function of this receptive field is analogous to taking a 2-dimensional convolution over the visual scenes with the difference-of-Gaussian (DoG) filter, such that the contrast between each small area of a visual scene and its surrounding is enhanced. 

Interestingly, the arrangement and projection of both RHCs and BPs are similar among many different species. There are only subtle differences; for example, RHCs with different dynamics and projection ranges exist in different animals and may be beneficial in various environments \cite{hctype}. Is this conservativeness due to the fact that this general pattern is the best animals could have? Though cannot answer the open question directly, we hypothesize that there are alternative structures that lead to local maxima of fitness, but do not exist or are rare in real world for unaddressed reasons. To explore those alternatives, genetic algorithm (GA), a random-based searching algorithm optimal for identifying multiple local extreme values in a fitness surface \cite{ga}, is used to simulate the evolutionary process and analysis of the resulting structures with high fitness, or fitness scores, is performed, in order to understand how they work and how noise level and evaluation tasks could affect their formation.

\section{Method}

\subsection{Visual Input}
In the interest of efficient model development, inputs are 1-D; the anatomy of the phenotype is hence reduced to essentially 2-D structures.  The level of stimulation to a rod cell by a certain spectrum of light is represented by a scalar. Under this simplification, we make reference signals for the ideal scenario, where only an edge of the target is visible in this small portion of retina, and slightly perturbed the signals to mimic interventions such as dusts, water, grass, and other common natural objects. A reference signal is thus a step input (Fig. 2a), where the edge locations are randomly generated. A perturbed signal is generated by firstly introducing uniform noise to the reference signal, and then convolving with an appropriate Gaussian filter to eliminate superfluous serration (Fig. 2a).

\subsection{Representation of Retina}
Though retinas are commonly modeled simple convolutions with DoG \cite{dog4, dog3, dog2, dog1}, this approach circumscribes efficient alternation of neuronal properties, and does not support complex structures, such as feedback and recurrent connections, which could lead to more complex but effective receptive fields. We therefore take the morphological approach, neural network.

We assume that all retina neural networks to be constructed are taken from parts of the whole retinas that have the same area and spacing of photoreceptors and RGCs\footnote{We use "retina(s)" as the alias of the partial neural networks in the rest of the text, unless the term "whole retina" is used to refer to the large, integral 2-dimensional retina of an eye.}. That is, the amount of photoreceptors and RGCs are the same for all neural networks; the variation in the number of photoreceptors and RGCs should also be disabled because the algorithm would otherwise maximize the number of receptors and RGCs to increase visual acuity, and leave the rest of processing to the higher processing unit (that is, the perceptron described below). Moreover, each neuronal type\footnote{To avoid the confusion between "type" and "layer", and to downplay the spatial order of different types of interneurons, the term "layer" is circumvented.} is assumed to uniformly tile the space as well, given the assumption that only a small portion of a retina is taken, as well as for the sake of convenience. 

During simulation, the photoreceptors receive constant input continuously, and their states can be modulated by interneurons. Interneurons can receive input from the photoreceptors (Fig. 2a). The only spiking neurons in the retina, RGCs receive direct input from the photoreceptors and processed input from other interneurons, and fire action potentials. The exponential integration-and-fire model (EIF) is used to simulate the spiking activities of RGCs. The general retina processing can be described as a set of differential equations:

\begin{align}
    \tau^{(0)}\frac {\mathrm{d}} {\mathrm{d} t} V^{(0)}(t) &= -(V^{(0)} - V_{rest}) + I_{ext} + I_{int}^{(0)} + I_{lim}^{(0)} \\
    \tau^{(j)}\frac {\mathrm{d}} {\mathrm{d} t} V^{(j)}(t) &= -(V^{(j)} - V_{rest}) + I_{int}^{(j)} + I_{lim}^{(j)},\hspace{5mm} j \notin \{0, g\} \\
    \begin{split}
    \tau^{(g)}\frac {\mathrm{d}} {\mathrm{d} t} V_k^{(g)}(t) {} &= -(V_k^{(g)} - V_{rest}) + \Delta e^{(V_k^{(g)}-\theta) / \Delta} \\
    &+ I_{int, k}^{(g)} + I_{lim}^{(g)} + I_{reset, k}(t)
    \end{split}
\end{align}

Here capitalized symbols denote vectors or matrices. $V$ is the membrane potentials of a type of cells, a time-dependent variable with time constant $\tau$ and resting potential $V_{rest}$. In our model, $V_{rest} = 0.5$. Received only by photoreceptors, $I_{ext}$ is the external input or the visual signal (Eq. 1). $I_{int}$ is the internal input or the weighted input from the other neurons, and $I_{reset, k}$ is the reset current of the $k$-th RGC immediately after it fires at time $t_f$ (Eq. 2, 3, 4, 5). The spike train $S_k(t)$ is represented as the sum of Dirac delta functions at different times (Eq. 6).
\begin{align}
    I_{int}^{(j)} &= \sum_{i \in C^{(j)}} V^{(i)^T} W_{ij}\\
    I_{reset, k}(x) &= -\tau^{(g)} (\theta_r - V_{reset}) S_k(t)\\
    S_k(t) &= \sum_{t_f} \delta(t - t_f)
\end{align}
where $C_j$ (\textit{section 2.4.1}) is the set of types to which the neurons of type $j$ project (Eq. 4). 

In addition, to make the membrane potential in a proper range (here the range is set to $[0, 1]$), a limitation current $I_{lim}$ is introduced (Eq. 1, 2, 3). $I_{lim}$ in our model simply set the derivative to the corresponding boundary if the membrane potential exceeds $[0, 1]$ for non-spiking neurons and $[0, \infty)$ for RGCs. It is possible and more biologically plausible to use a nonlinear continuous function or model the adaptation as a dynamic system, but for our focus they do not make significant different in terms of results.
Finally, the average firing rates $F$ of RGCs are calculated and serve as the input to the trainable perceptron (Eq. 7 \& Fig. 2a; the perceptron will be described in \textit{section 2.4.5}).
\begin{align}
    F_k = \frac 1 T \int_T S_k(t) \mathrm{d} t
\end{align}

\begin{figure}
    \centering
        \includegraphics{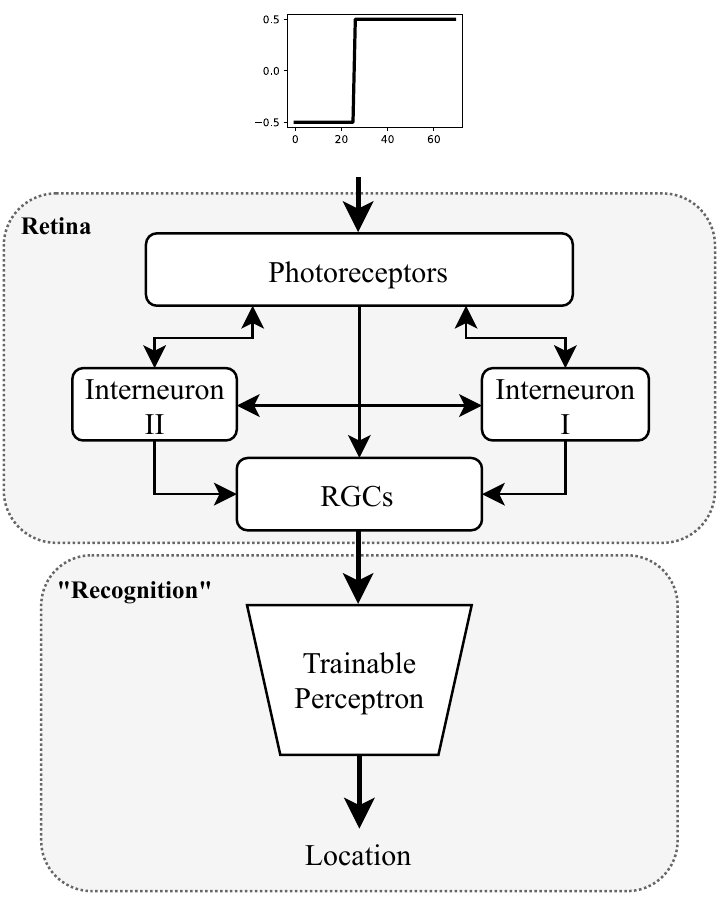}\\(a)\vspace{3mm}\\
        \includegraphics{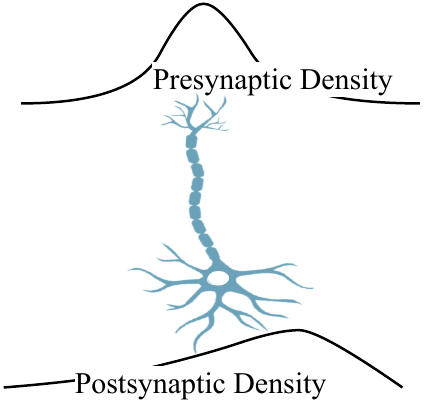}\\(b)
    \caption{
    (a) A diagram of the perception with a retina. The generated signal is given to the retina with photoreceptors, interacting interneurons and spiking RGCs. The interneurons interact among different types (solid bidirectional arrows). Only 2 types are drawn to save space. The firing rates of RGCs is the data for the trainable perceptron, re-initialized at every generation, to identify the locations of the edges.
    (b) The pre- and postsynaptic spatial density distributions of a neuron. Because the thickness of the retina is assumed to be negligible, each distribution is 1-dimensional, along the lateral direction. We choose Beta distribution for its variable shapes.
    }
\end{figure}

\subsection{Genetic Algorithm}

GA is advantageous in this task because it does not require gradient while it is hard to use gradient descent for the retina structures. Moreover, the algorithm, depending on the implementation, can be very fast and allows parallelism.

\subsubsection{Genome}
As described, each retina phenotype is built from the corresponding chromosomes. Each chromosome is composed of a set of alleles, beginning with an integer, $n_t$, which specifies the number of interneuron types, and is upper-bounded by a constant $N_t$. The initial $n_t$ for all individuals is 0, i.e. no interneurons. In the following generations, the individuals can gain interneurons through mutation and crossover (\textit{section 2.4.3, 2.4.4}). Each neuronal type, $i$, including the photoreceptor and RGC, has 
\begin{enumerate}
    \item The number of photoreceptors is fixed at a value $N_c$ for all chromosomes.  A number of cells $n_c^{(i)}$ denoting how many cells of interneuron type $i$ in the selected region of a large retina (\textit{section 2.1}), with a constant "biological upper limit" $N_c$.  The number of RGCs at has a constant value $N_g$, for all chromosomes. 
    \item Binary encoding of pre- and postsynaptic features, $x^{(i)}_a$ and $x^{(i)}_d$ respectively, where subscript $a$ stands for axon and $d$ for dendrite. The encoding is abstract and inspired by Hopfield network, and the sign of connections between two types $i$ and $j$, i.e. the polarity, is determined by the product of the pre- and postsynaptic feature encoding (Eq. 9).
    \item Shape parameters, $\alpha^{(i)}_a$, $\beta^{(i)}_a$, $\alpha^{(i)}_d$ and $\beta^{(i)}_d$, for axon and dendrite spatial density distributions. In our model, cells are assumed to uniformly distributed in the space with width 1, and the thickness of the retina is assumed to be negligible, so the distributions can be 1-dimensional distributions and the input is defined as the lateral distance between two neurons. Here we use Beta distribution for its variable shapes (Fig. 2b). The domain for the parameters is a small subset $B \subset [0, \infty)$, so as to enhance the run time by reducing nuances between the values. 
    \item A time constant $\tau^{(i)}$ that controls the rates of change of neurons' internal states (Eq. 1-3). Its value is sampled from an empirical range $[5, 100]$ with sampling interval 5. 
    \item A set $C^{(i)}$ containing the indices of the postsynaptic types in retina that this type of neurons projects to (Eq. 4). For RGCs, this set is always empty.
\end{enumerate} 

\begin{table*}
    \centering
    \caption{Definitions and ranges of variables that evolve with generations in the simulation. All superscripts "$(i)$" denote an arbitrary neuronal type $i$. $N_t$, $N_c$, and $N_g$ are constants, representing the maximum numbers of interneuron types, non-RGC cells, and RGC cells respectively. $A = \{N_c\}$ for photoreceptor, $A = [1, N_c]$ for interneuron type, and $A = \{N_g\}$ for RGC.}
    \label{tab:def}
    \begin{tabular}{lll}
        \toprule
        Variable & Definition & Domain\\[10pt]
        \midrule
        $n_t$ & Number of interneuron types & $[0, N_t] \cap Z_+$\\
        $n_c^{(i)}$ & Number of cells of type $i$ & $A_i \cap Z_+$\\
        $x^{(i)}_a, x^{(i)}_d$ & Encoding of pre- and postsynaptic terminal "polarity" & \{-1, 1\}\\
        $\alpha^{(i)}_a, \beta^{(i)}_a$ & shape parameters for axon density distribution & $B \subseteq [0, \infty)$\\
        $\alpha^{(i)}_d, \beta^{(i)}_d$ & shape parameters for dendrite density distribution & $B \subseteq [0, \infty)$\\
        $\tau^{(i)}$ & Time constant & $[5, 100]$\\
        $C^{(i)}$ & Set of postsynaptic types & \\
        \bottomrule
    \end{tabular}
\end{table*}

All genes of each retina are randomly initialized with valid values in their domains, except for the number of interneuron types. Considering the initial generation as the very early stage of retina evolution, we set the initial number of interneuron types to zero. This gene, described further in \textit{section 2.4.4}, can change as the result of duplication or deletion of interneuron types. 

From the genome, the connection $W^{(i,j)}$ from neuronal type $i$ to $j$, if $j$ in $C^{(i)}$, is the product between the polarity, $p^{(i)}$, and the connection matrix from a generator that utilizes the lateral distances $D^{(i,j)}$ between the two types, the axon density distribution of $i$, and dendrite density distribution of $j$. 
\begin{align}
    W^{(i,j)} &= p^{(i,j)} f(D^{(i,j)})\\
    p^{(i,j)} &= x^{(i)}_a x^{(j)}_d\\
    f(D^{(i,j)}) &=  cH(P(D^{(i,j)} | \alpha^{(i)}_a, \beta^{(i)}_a) \times P(D^{(i,j)} | \alpha^{(j)}_d, \beta^{(j)}_d))
\end{align}
where $c$ is a constant that scales the weights. Here, we simply take the advantage of the shape of the Beta distribution by clipping the joint distribution using a Heaviside function $H$ (Eq. 10), but a stochastic generator is also feasible. Both approaches are tested to be effective, but the nonstochastic one is used for better accountability. Table \ref{tab:def} contains the definitions of genes and their ranges.

\subsubsection{Selection}
A fraction of the population with the greatest fitness, called elites, $P_e$, are preserved, and the others are or may be replaced by "children," so that after one generation, some good results that the simulation achieves so far would not be removed. It is important to set the fraction of elites appropriately, as too large $P_e$ would lead the optimization converge to local minima early, i.e. premature convergence, while too small $P_e$ would introduce too much randomness and impede the optimization. There is no consensus on the number of elites, and is believed to be task-specific. We choose the value to be between 5\% and 10\%.

The remainder of the succeeding generation is populated by applying the binary tournament selection algorithm \cite{Miller95geneticalgorithms}, in which two rival parent chromosomes are drawn from the population without replacement (Fig. 3a). The rival with better fitness evaluation has higher chance ($> 0.5$) to be one of the parents, while the one with worse fitness evaluation has lower chance to be the parent. the second parent is chosen the same way. Here, unlike the conventional approach where the chance is fixed, the selection probability depends on the ratio of the two fitness scores, $s_1, s_2$ (Eq. 11). If two rivals are nearly equally competent, the ratio between their scores is close to 1, and the chances for both are close to 0.5. In case of comparably good retinas, this dynamical procedure potentially prevents premature convergence.
\begin{align}
    p_i &= \frac{s_i}{s_i + s_j}
\end{align}

where $s$ is defined as the $R^2$ score (\textit{section 2.4}).

\subsubsection{Crossover}
\begin{figure}
    \centering
    \includegraphics{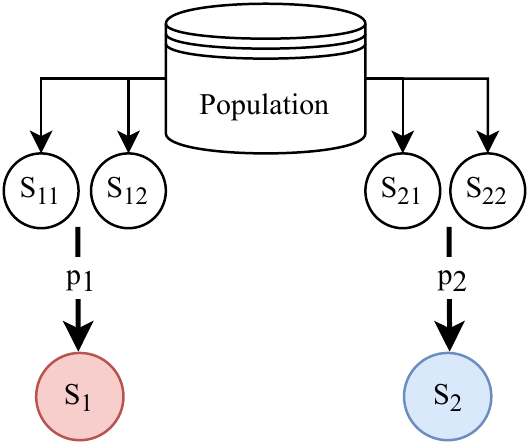}\\ (a) \vspace{3mm}\\
    \includegraphics{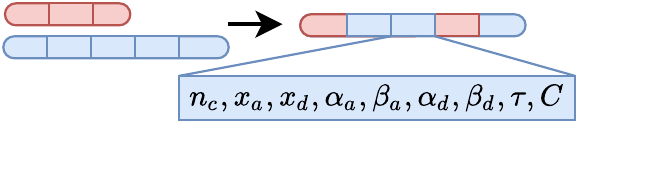}\\ (b)
    \caption{
    (a) Selection. Two couples of rivals are drawn from the pool, and the selection of each "parent" is a Bernoulli process with probability determined by Eq. 11.
    (b) Crossover. The colors of the "chromosomes" are coherent with Fig. 3b. Each segment of a "chromosome" is a set of "linked genes" of a neuronal type, i.e. set of variables determining the properties of a neuronal type that can only be exchanged as a whole. }
\end{figure}
Unrestricted crossover is observed to bring up variation during each generation that possibly leads to either premature convergence or divergence. So as to limit variability, the properties that are type-specific are linked, so that either none or all of them are passed to an offspring (Fig. 3b). Moreover, the crossover is a Bernoulli process, such that non-elites have the chance to survive to the next generation, even though they still have to go over mutation. In this way, the randomness is kept at a moderate level, and some "advantageous parts" of the non-elites may be preserved. 

During crossover, a "child" inherits $n_t$ from one of the "parents" randomly. Since photoreceptor and RGC are special types that cannot interchange with interneuron types, the "child" also randomly gets the properties of them from one parent. Then, each interneuron types of a "child" is inherited from one of the "parents." Note that the crossover of interneuron types does not require any correspondence in their type indices. For example, the 1st interneuron type of a "child" can come from the 5th interneuron type of its "parent."

\subsubsection{Mutation}
To avoid non-convergence due to high randomness, the chances to mutate a type and a gene within the type are not high; the amount of mutation for each gene is kept small as well. Since all genes, except for $C^{(i)}$, are either discrete or sampled with some intervals, the mutation to each gene can be viewed as a random walk whose step size follows a Poisson process, with constraint that a value is kept at its nearest boundary if the random walk can cause the value to be out of range. For $C^{(i)}$, in each generation there is a small chance to add an element to or delete one from the set. 

In each generation, each individual also has a small chance to duplicate or delete one type, if the resulting number of interneuron types is not out of range, and if neither the receptor nor RGC type is deleted. If either duplication or deletion occurs, the indices of neuron types are updated and each type's set of postsynaptic types will be updated. For duplication, the genes are subject to mutation in the same generation.

\subsection{Fitness Evaluation}
Defining an evaluation metric as all-encompassing as natural selection is hard and may even be unfeasible. To simplify the problem, we postulate that, in nature, the effectiveness of primary retina processing forms the major selective pressure. Even though the number of synapses is also important as it is related to energy consumption and light absorbance \cite{synapseenergy}, no conclusion is made in previous studies about how this factor can influence survival. We only introduce a shape gain with a small weight, so as to penalize those with complex, intertwining connections that however, compared to simpler alternatives, contribute trivially by increasing this complexity. The shape gain is given by
\begin{align}
    g = \left\{\begin{matrix}
    0 & C^{(0)} = \varnothing \text{ or } n_t = 0\\
    \sum_{i = 1}^{n_t}\frac{1 - |L(C^{(i)}) - 1|}{n_t} & \text{else}
    \end{matrix}\right.
\end{align}
where $L(C)$ is the size of the set $C$. Structures where interneurons are connected to more types will thus have smaller gains. 

\begin{figure}
    \centering
    \includegraphics[width=3.2in]{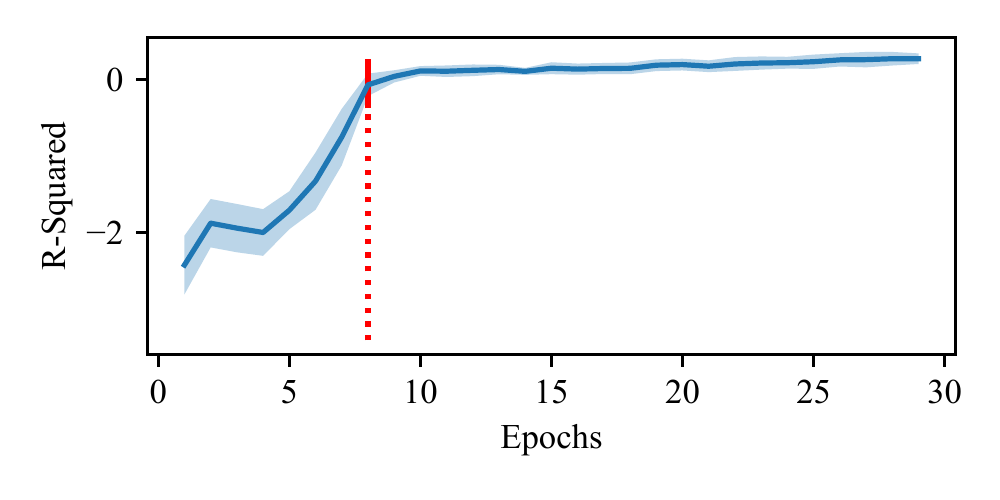}
    \caption{The $R^2$ scores of the perceptrons on testing set ($n = 30$) for different training epochs. The blue shaded area represent $\pm 1$ standard deviation around the mean (blue solid curve). Here, the $R^2$ scores for up to 30 epochs are showed, because the $R^2$ scores in the next 70 epochs are nearly the same as those after 20 epochs. The elbow point occurs at 8 epochs (red dash line).}
\end{figure}

The effectiveness of the primary retina process of a retina is evaluated with a perceptron re-initialized at every generation, which represents early simple cognitive processing. The perceptron has to perform regression on the location of change of each step input. The evaluation is a measure of how well the retina can emphasize the boundaries of the input signals, while not restricting how the retina should process the input, or producing certain patterns of firing; it is comparable to the inference in nature, where visual animals need to infer where an object's boundaries locate.  

At each generation, every retina receives a number of input (\textit{section 2.2}) generated randomly at the beginning of the whole simulation. The outputs from the RGCs are used to train a perceptron, which is initialized at a fixed constant for each retina at every generation, and then trained with batch gradient descent. The sample size, learning rate, and the number of training epochs are set based on preliminary experiments, such that they are insufficient for the perceptron to be well-trained or overtrained; that is, the retina should preprocess the raw inputs to some degree that makes the perceptron easier to train. In testing, another fixed set of input is given to each retina, whose output is used to test its trained perceptron. The fitness score of a retina $s$ is the $R^2$ score of the performance of its perceptron, i.e. $1 - SS_{residual} / SS_{total}$, with a fixed range $[-1, 1]$ for easy comparison.

\subsection{Weak Perceptron}
To decide the sample size, learning rate, and the number of training epochs that make a weak perceptron, multiple perceptrons are trained with variable number of visual samples, i.e. the input to retina, and learning rate. The other hyperparameters are fixed at empirically values, and the perceptrons are trained in 100 epochs. We find that 500 samples and learning rate of 0.01 are appropriate, considering the run time. With the sample size and learning rate, we then train the perceptron for 30 trials, each consisting of the training with maximum number of epochs from 1 to 100, and calculated the $R^2$ scores of the perceptrons on the test sets (Fig. 4). The $R^2$ scores for each number of epochs are roughly normally distributed. The mean $R^2$ at the plateau is close to 0, indicating that with our hyperparameters, the perceptrons cannot do better than guessing; the retina is thus important in preprocessing. We find the elbow point is at Epoch 8, and use 7 as the maximum number of epochs for our model to introduce extra difficulty. Thus, a good retina structure should, intuitively, not only raise the plateau of this curve, but also make it appear at least one epoch earlier.

\section{Results}
\subsection{GA Optimizes the Retina Structures of the Elites}
Multiple trials of 400 generations have been run and the elites at the final generation are collected for analysis. In each generation, the population size is 150 and the elite size is 10. During mutation, both of the chances to skip a type and to skip one single gene are 0.5, while the probabilities of duplication and deletion are both 0.3. The crossover rate is 0.2. The weight of $R^2$ is 0.85 and that of shape gain is 0.15.

Across all trials, the survival gain, defined as the difference between the minima of the elites at the first and last generations, is significantly greater than 0 (left-tailed p test, $p = 9.3 \times 10^{-44} < 10^{-3}$). The survival gain together with the non-decreasing fitness scores demonstrates the efficacy of our algorithmic design that it optimizes the retina structures of the elites. Note that the “evolution” does not happen in every generation, but manifests several breakthrough points, between which the intervals, or stasis, vary (Fig. 5, top). During each elite's stasis, its genome does not vary.

In comparison, the medians and IQRs of the populations do not change significantly across all trials (Fig. 5, bottom). That is in our expectation, because our algorithmic design neither boosts nor suppresses the survivals of the non-elites. \textit{section 4}
will discuss the implication of elites and non-elites in GA with respect to natural evolution.

\begin{figure}
    \centering
    \includegraphics[width=3.3in]{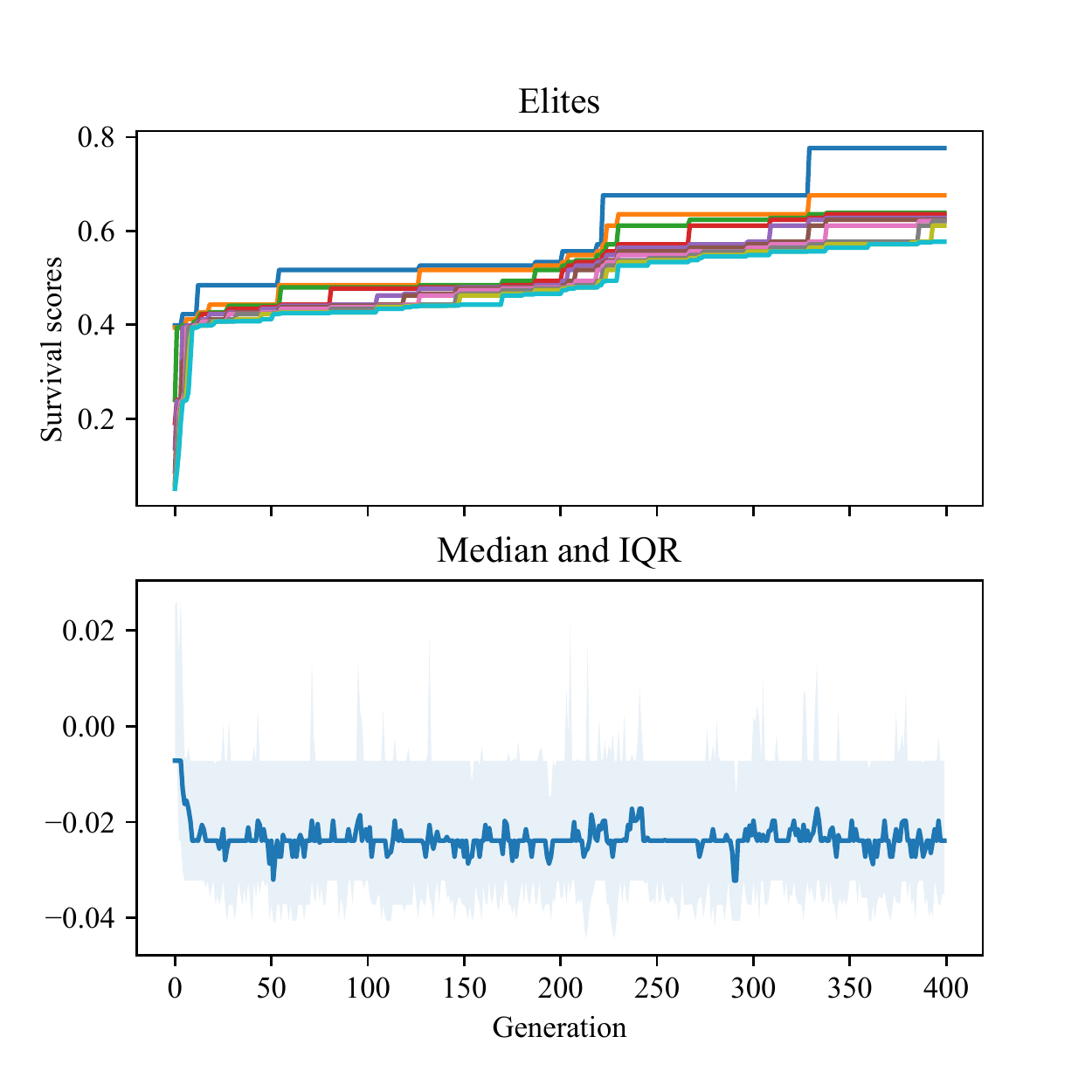}
    \caption{An example of simulation. The fitness scores are defined as the weighted sum between $R^2$ scores and the shape gain described above. Top row: the fitness scores of elites at different ranks (different colors) over generations. Bottom row: the medians (blue solid curve) and IQRs (blue shaded area) of all individuals over generations. }
\end{figure}

\begin{figure}
    \centering
    \begin{tabular}{cc}
        \includegraphics[width=0.21\textwidth]{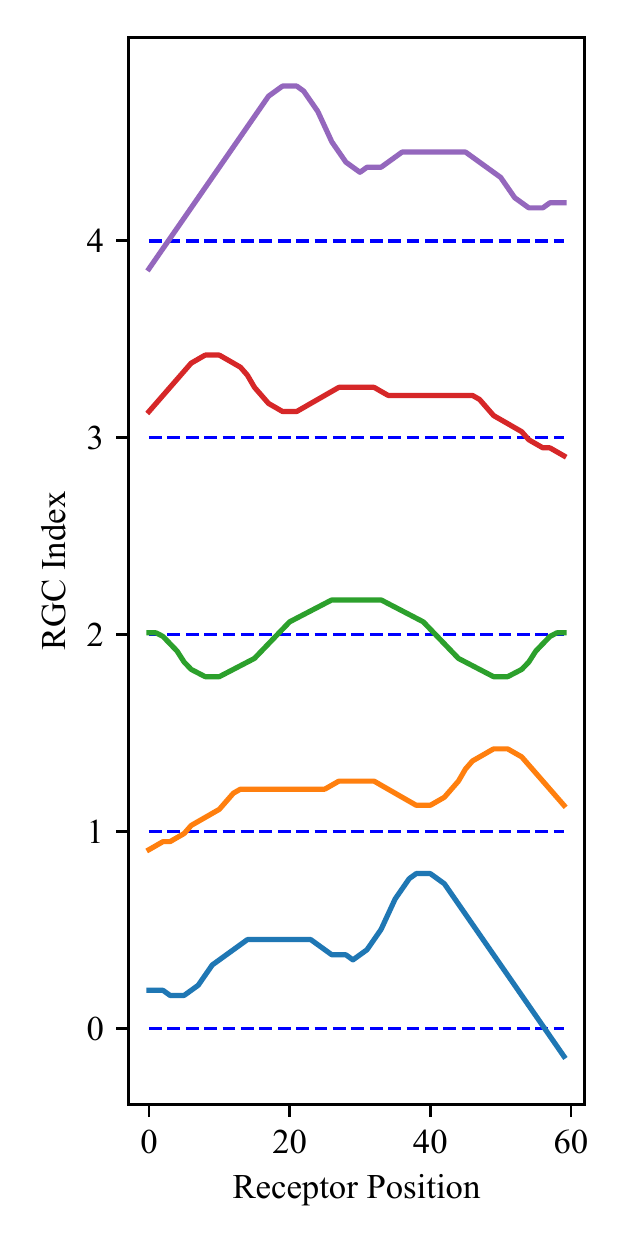}
        &
        \includegraphics[width=0.21\textwidth]{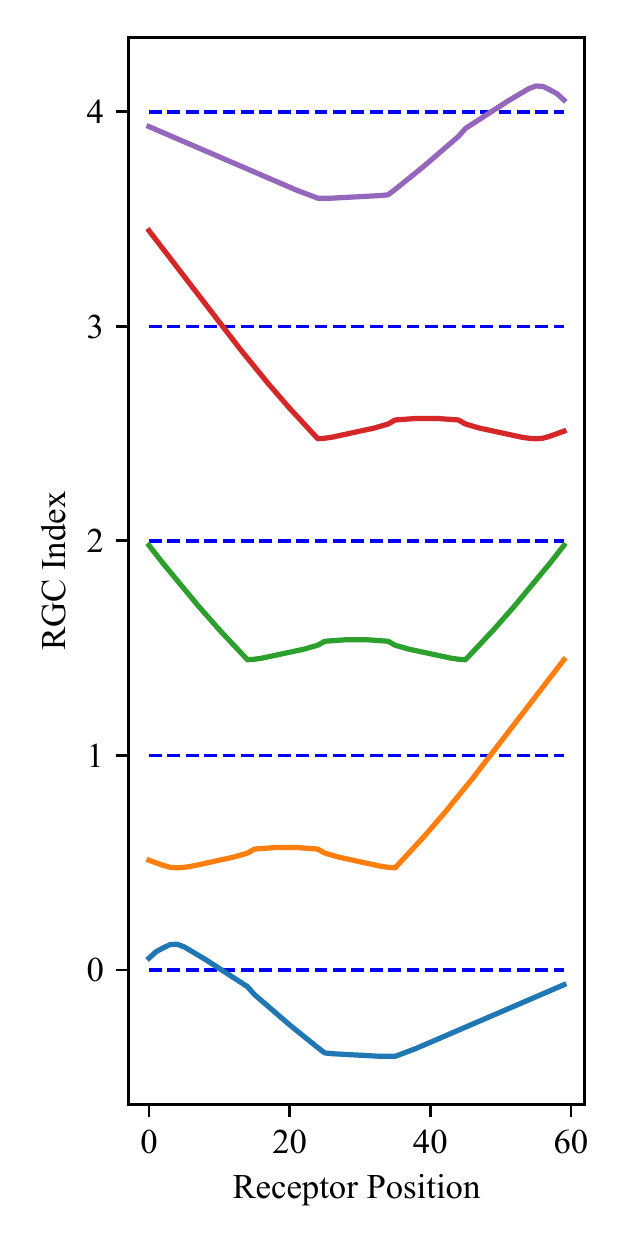}\\
        (a) & (b)
    \end{tabular}
    \caption{The best elites from two arbitrarily selected trials. In each subfigure, each row represents the spatial tuning curve of one of the five RGCs, with respect to the photoreceptor index. The blue dotted line in each row stands for the baseline firing.}
\end{figure}

\subsection{Center-Surround Receptive Fields in the Final Elites}
The elites with the best fitness scores always appear at the few generations, because the scores are non-decreasing. Case studies of the connection matrices and tuning curves of the elites at the final generations from different trials are performed. A few elites' central RGC has center-surround receptive field resembling that of biological retina (Fig. 6a), while some elites' lateral RGCs could have such receptive field, if their tuning curves are symmetrical at their locations (Fig. 6a). The other elites have more complex tuning curves. However, the most important trait is that all elites’ RGCs have well separated spatial tuning curves, such that the firing patterns produced are more informative and allow the perceptrons to learn better.

\section{Discussion}
\subsection{Similarity to the Punctuated Equilibrium Theory}
The evolution of our model, as measured by the most fit individual in each generation, is not gradual, but progresses in abrupt jumps.  The "steps" in the improvement of the elites occur at variable intervals. The plateaus correspond to "stasis" in the theory of punctuated equilibrium \cite{punc_equ}; during these intervals, the retina topologies remain unchanged.  A morphological change  (corresponding to the emergence of a new "species"), can result in a step in the fitness function.  

In genetic algorithm, the elites correspond to the primary existing species, while the more extreme mutations generate new retinal topologies, some of which are "fatal" mutations, and occasionally are improvements.  In nature, these "attempts" happen less often over the course of many generations and in smaller scale with each generation; while in genetic algorithm, the frequency and scale are drastically magnified. The unsuccessful mutations in genetic algorithm, whose failure, in our model, is defined as not being able to become an elite (or, analogously, become one species), disappear in a few generations like those in nature. The successful, new species emerge with notable changes in genomes when they first appear, which are reflected by the fitness scores in our simulation; the existing species, if still successful in the competition, keep existing for multiple generations with little change (Fig. 5, top; note that the curves are colored by the ranks and the same species are the horizontally aligned line segments). This phenomenon is consistent with the punctuated equilibrium theory. 

However, note that, unlike our simulation, in nature a new species may not necessarily have higher "fitness scores" than their antecedents; factors such as genetic drift may give rise to a more complex, non-monotonic relation between changes and survival. Moreover, natural environment changes over time, so the monotonicity of survival over generations in our simulation may not hold for natural evolution. Here, we also emphasize that the "generation" we use is an abstract, dynamic duration, not a fixed period of time.

\subsection{Factors to be Considered and Limitation of Computational Simulation}
Our model contains a simple approach to evaluate the fitness. However, for more realistic modeling of the evolution, it is critical to find more properly complex fitness evaluation for GA, which is especially hard, for evolution requires more degrees of freedom and involves numerous factors. Other approaches such as measuring the sharpness of a signal are potentially effective, though finding the way of measuring such sharpness is critical. A measure of using fast Fourier transformation can be suitable for this simulation from some simple experiments conducted \cite{fft}. However, it is hard to tell if assessing the sharpness is the only way to preprocess the input for later perception and learning. On the other hand, the perceptrons can be substituted by deeper neural networks and reinforcement learning models, which enable more complex inputs and retina neural networks but also drastically increase the computation. The reliability of the complex substitutions also necessitates careful evaluation. 

Selective pressure is implemented by the definition of a fitness function, which is much more complex in nature than the one we have defined for the model. At microscopic level, the possibilities for channels, neurotransmitters, supportive proteins \cite{evo_lamb}, etc. to appear in the genome and be expressed in the eyes may vary during the natural evolution; but they are out of scope for this paper and not fully elucidated in evolution research. Retinal neurogenesis, not incorporated in our model, also plays an important role. Other physical factors, like light absorbance by the neural fibers and cellular energy consumption, can affect evolution as well. At the input level, it is well believed that the eye structures not only need to support primary visual processing, but also proffer other functionalities, such as protection, or are subject to other selective pressure, such as volume. Whether these factors affect the retina network organization is unclear. Finally, at behavioral, social, and environmental levels, multiple factors such as mating and bottleneck effect can exert larger-scale changes in the genome, whose influence is hard to disentangle for retina only.

There are also limitations in our computational simulation. Obviously, the computational power may not be sufficient to contain all factors described above. In terms of our model, one significant limitation is that the local connections of the neurons that are related to the boundaries, and thus the portions of tuning curves near the boundaries, may be affected by the insufficient amount of input. That partially explains why the tuning curves are quite different from each other in the results. Yet, it is not the case if the retina portion is "put back" to the large one (as described in \textit{section 2.2}, we consider our small neural network as a tiny portion of a large retina). Another limitation is that we neither incorporate synaptic plasticity nor dynamically scale the weights, so some interneuron types may be added solely for altering the RGC firing generally---a process that can be more naturally encoded by having a proper adaptation mechanism.

By examining connection matrices, we do not observe explicit lateral inhibition connections facilitated by horizontal cells. In spite of the stochasticity of this searching process, it is possible that the simple regression task does not necessitate the lateral inhibition structure in biological retina, which provides more complex functional supports to higher level processing. For example, biological retina needs to be compatible with functions such as motion perception, while not all shapes of tuning curves would fulfill the requirement. The evolved one-dimensional retina structures with only a single receptor type may not support these functions.

Thus, it is expected that the found structures are not exactly the same as the real retina neural networks. However, our model yields neural networks that have DoG-like turning curves, and, for the effective alternatives, well separated tuning curves. Our method can be furthermore refined, such as substituting the perceptrons with more complex, biologically inspired neural network models, and taking more factors into account. It is possible that adding more details will either cause convergence into a single structure, or let people to discover more interesting, effective structures. 

\subsection{Implications for Goal-driven Search and Sustainable Enhancement of Neural Network Models in Machine Learning}
The method we use, and the analysis conducted, could provide insight into meta-learning, i.e. learning the hyperparameters of deep learning models. One might be able to define the selective pressure that conveys the key requirements of the task, so as to find several good structures of the model to start with. Although currently there are other hyperparameter-tuning schemes, empiricism is still not uncommon in determining hyperparameters of deep artificial neural networks (DNNs). This framework, which considers biological plausibility, encodes the network to decrease parameter size, and yields the “big pictures” of network topology that people can possibly adopt to devise and train with gradient methods, can be used in tuning the hyperparameters of DNNs with less empiricism involved, and is therefore beneficiary to those who are not very experienced in deep learning. 

Moreover, simulated evolution has the potential to further improve deep neural networks and artificial intelligence (AI) models \cite{eaai1, eaai2, eaai3, eaai4}. This nature-inspired approach can also provide inspiration for the artificial evolution of AI. Some of these have been attempts on combining artificial neural networks (ANNs) and evolutionary algorithms, termed as topology and weight evolving ANNs (TWEANNs). However, many of these attempts only focus on adapting evolutionary algorithms to alter the node lists and connections directly, possibly leading to tremendous computational cost for DNNs. More importantly, these algorithms discard the gradient descent component completely. 

We thus raise the possibility that a deep neural network can be optimized by combining evolution frameworks and gradient descent: the former on the topology and the latter on the weights. In our framework, the retina is analogous to a part of a large network whose topology needs to be optimized, and the trainable perceptron corresponds to the remaining part of the network. A possible scheme is that, in each iteration, a model for an image-related task can be separated into 
\begin{enumerate}
\item a structural evolving part, which is not too large and is improved using evolutionary frameworks, analogous to the retina in our model,
\item a fixed, trained part that provides other processing,
\item a trainable, evaluating part, such as the dense layers of a CNN, whose output can be used to compute the fitness scores of part (1), resembling the perceptron in our model. 
\end{enumerate}
The weights of the whole or part of the network can be trained again after the topology of part (1) is evolved. The new weights and topology will be used in the next iteration, where the same or different portion of the network will become the structural evolving part. 

The whole process is automatic, and since everything can be backed up, it is possible to use the model for real world tasks while improving the structures---like natural evolution---and switch to another branch whenever the improvement at the current branch fails. It is also advantageous that this paradigm can support transfer learning. On the other hand, it is possible to utilize these backups for bagging or random forests, potentially leading to more powerful architectures.

\printbibliography

@article{limulus,
    author  =   "H. K. Hartline and Floyd Ratliff",
    title   =   "Inhibitory Interaction of Receptor Units in the Eye of Limulus",
    journal =   "J Gen Physiol",
    volume  =   "40",
    number  =   "3",
    pages   =   "357--376",
    year    =   "1957",
    doi = "10.1085/jgp.40.3.357"
}

@article{lateralinhib,
    author  =   "Wallace B. Thoreson and Stuart C. Mangel",
    title   =   "Lateral interactions in the outer retina",
    journal =   "Prog Retin Eye Res",
    volume  =   "31",
    number  =   "5",
    pages   =   "407--441",
    year    =   "2012",
    doi = "10.1016/j.preteyeres.2012.04.003"
}

@article{hctype,
    author  =   "Serge Picaud and David Hicks and Valerie Forster and Jose Sahel and Henri Dreyfus",
    title   =   "Adult Human Retinal Neurons in Culture: Physiology of
                Horizontal Cells",
    journal =   "Invest Ophthalmol Vis Sci",
    volume  =   "39",
    number  =   "13",
    pages   =   "2637--2648",
    year    =   "1998"
}

@article{fft,
    author  =   "Kanjar De and V. Masilamani",
    title   =   "Image Sharpness Measure for Blurred Images in Frequency 
                Domain",
    journal =   "Procedia Engineering",
    volume  =   "64",
    pages   =   "149--158",
    year    =   "2013",
    doi = " 10.1016/j.proeng.2013.09.086"
}

@article{dog1,
    author  =   "Xian-Shi Zhang and Shao-Bing Gao and Chao-Yi Li and Yong-Jie Li",
    title   =   "A Retina Inspired Model for Enhancing Visibility of Hazy Images",
    journal =   "Front Comput Neurosci",
    volume  =   "9",
    number  =   "151",
    pages   =   "1-16",
    year    =   "2015",
    doi = " 10.3389/fncom.2015.00151"
}

@article{dog2,
    author  =   "Nikos Melanitis and Konstantina S. Nikita",
    title   =   "Biologically-inspired image processing in computational retina models",
    journal =   "Computers in Biology and Medicine",
    volume  =   "113",
    pages   =   "",
    year    =   "2019",
    doi = " 10.1016/j.compbiomed.2019.103399"
}

@article{dog3,
    author  =   "Yeonan-Kim J and Bertalmı´o M",
    title   =   "Retinal Lateral Inhibition Provides the Biological Basis of Long-Range Spatial Induction",
    journal =   "PLoS ONE",
    volume  =   "11",
    number  =   "12",
    pages   =   "1--23",
    year    =   "2016",
    doi = " 10.1371/journal.pone.0168963"
}

@article{dog4,
    author  =   "D. Marrand E. Hildreth",
    title   =   "Theory of Edge Detection",
    journal =   "Proc R Boc Lond",
    volume  =   "207",
    pages   =   "187--217",
    year    =   "1980"
}

@ARTICLE{Miller95geneticalgorithms,
    author = {Brad L. Miller and Brad L. Miller and David E. Goldberg and David E. Goldberg},
    title = {Genetic Algorithms, Tournament Selection, and the Effects of Noise},
    journal = {Complex Systems},
    year = {1995},
    volume = {9},
    pages = {193--212}
}

@article{synapseenergy,
    author = {Harris, Julia and Jolivet, Renaud and Attwell, David},
    year = {2012},
    
    pages = {762-77},
    title = {Synaptic Energy Use and Supply},
    volume = {75},
    journal = {Neuron},
    doi = {10.1016/j.neuron.2012.08.019}
}

@article{evo_lamb,
    author = {Trevor Lamb},
    year = {2013},
    pages = {52 -- 119},
    title = {Evolution of Phototransduction, Vertebrate Photoreceptors and Retina},
    volume = {36},
    journal = {Progress in retinal and eye research},
    doi = {10.1016/j.preteyeres.2013.06.001}
}

@book{sensation,
    author  =   "Steven Yantis", 
    title   =   "Sensation and Perception",
    year    =   "2014",
    address =   "New York, NY",
    publisher = "Worth Publishers"
}

@book{ga,
    author  =   "Mitchell Melanie", 
    title   =   "An Introduction to Genetic Algorithms",
    year    =   "1999",
    address =   "Cambridge, MA",
    publisher = "MIT Press"
}

@INPROCEEDINGS{eaai1,
  author={P. A. {Vikhar}},
  booktitle={2016 International Conference on Global Trends in Signal Processing, Information Computing and Communication (ICGTSPICC)}, 
  title={Evolutionary algorithms: A critical review and its future prospects}, 
  year={2016},
  pages={261-265},
  address={Jalgaon},
  publisher={IEEE}
}

@Inbook{eaai2,
author="Gupta, Neeraj
and Khosravy, Mahdi
and Patel, Nilesh
and Gupta, Saurabh
and Varshney, Gazal",
title="Evolutionary Artificial Neural Networks: Comparative Study on State-of-the-Art Optimizers",
bookTitle="Frontier Applications of Nature Inspired Computation",
year="2020",
publisher="Springer Singapore",
address="Singapore",
pages="302--318"
}

@ARTICLE{eaai3,
  author={S. {Risi} and J. {Togelius}},
  journal={IEEE Transactions on Computational Intelligence and AI in Games}, 
  title={Neuroevolution in Games: State of the Art and Open Challenges}, 
  year={2017},
  volume={9},
  number={1},
  pages={25-41},
  doi={10.1109/TCIAIG.2015.2494596}
}

@article{eaai4,
author = "Panel Husbands and Inman Harvey and Dave Cliff and Geoffrey Miller",
title = "Artificial Evolution: A New Path for Artificial Intelligence?",
journal = "Brain and Cognition",
volume = "34",
number = "1",
pages = "130 - 159",
year = "1997",
issn = "0278-2626",
doi = "https://doi.org/10.1006/brcg.1997.0910",
url = "http://www.sciencedirect.com/science/article/pii/S0278262697909106"
}

@inbook{punc_equ,
author = {Eldredge, Niles and Gould, S.},
year = {1971},
month = {11},
pages = {82-115},
title = {Punctuated Equilibria: An Alternative to Phyletic Gradualism},
volume = {82},
booktitle = {Models in Paleobiology},
publisher = {Freeman Cooper \& Co},
address = {San Francisco},
}
\end{document}